\documentclass[runningheads]{llncs}

\usepackage{eccv}

\usepackage{eccvabbrv}

\usepackage{graphicx}
\usepackage{booktabs}

\usepackage[accsupp]{axessibility}  

\usepackage[pagebackref,breaklinks,colorlinks,citecolor=eccvblue]{hyperref}

\usepackage{orcidlink}

\renewcommand{\paragraph}[1]{\vspace{0.5em}\noindent\textbf{#1}}

\begin{document}

\title{NeRF-Insert: Local 3D Editing \\with Multimodal Control Signals} 

\titlerunning{NeRF-Insert}

\author{Benet Oriol Sabat\inst{1}\thanks{Work done during an internship at AWS AI Labs.},
Alessandro Achille\inst{2},
Matthew Trager\inst{2},
Stefano Soatto\inst{2}
}

\authorrunning{B.~Oriol et al.}

\institute{UCLA\\
\and
AWS AI Labs\\
\email{benet@cs.ucla.edu}, \email{\{aachille,mttrager,soattos\}@amazon.com}}

\maketitle
\begin{figure}[h]
\centering
\includegraphics[width=0.95\linewidth]{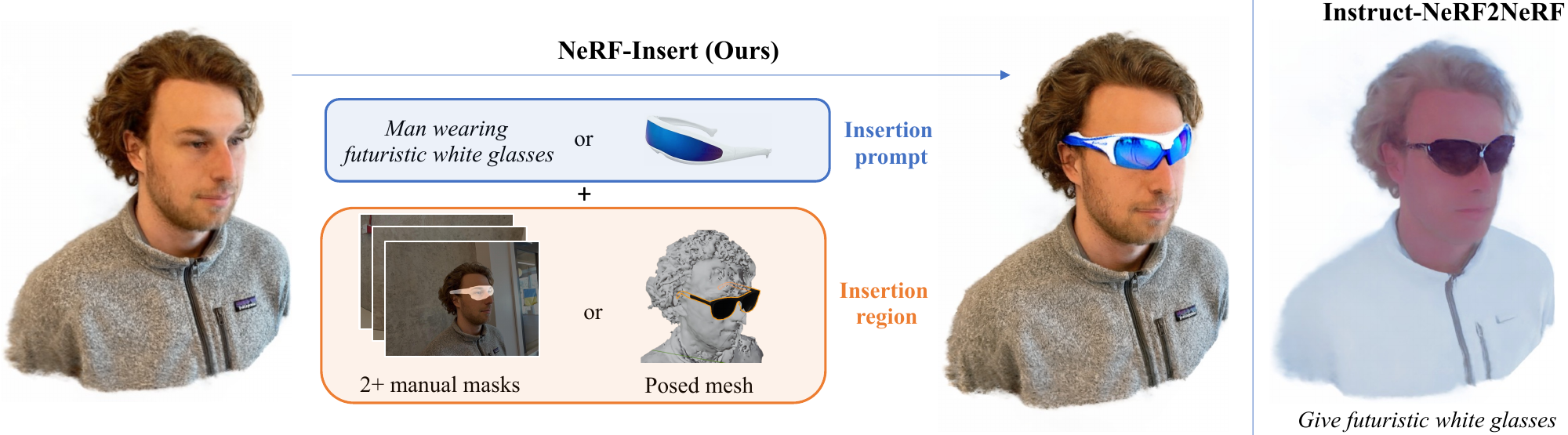} 
\caption{NeRF-Insert is a flexible framework for NeRF inpainting with different control modalities. A user can specify a 3D region with two or more manually-drawn image masks or by positioning a mesh/CAD model on the scene. Moreover, inpainting can be controlled with a textual prompt or with a reference image that influences the appearance of the inserted object or edited region.}
\label{fig:overall}
\end{figure}
\begin{abstract}
  We propose NeRF-Insert, a NeRF editing framework that allows users to make high-quality local edits with a flexible level of control. Unlike previous work that relied on image-to-image models, we cast scene editing as an in-painting problem, which encourages the global structure of the scene to be preserved. Moreover, while most existing methods use only textual prompts to condition edits, our framework accepts a combination of inputs of different modalities as reference.  More precisely, a user may provide a combination of textual and visual inputs including images, CAD models, and binary image masks for specifying a 3D region.  We use generic image generation models to in-paint the scene from multiple viewpoints, and lift the local edits to a 3D-consistent NeRF edit. Compared to previous methods, our results show better visual quality and also maintain stronger consistency with the original NeRF.
\end{abstract}

\section{Introduction}

\begin{figure}[h]
\centering
\includegraphics[width=0.99\linewidth]{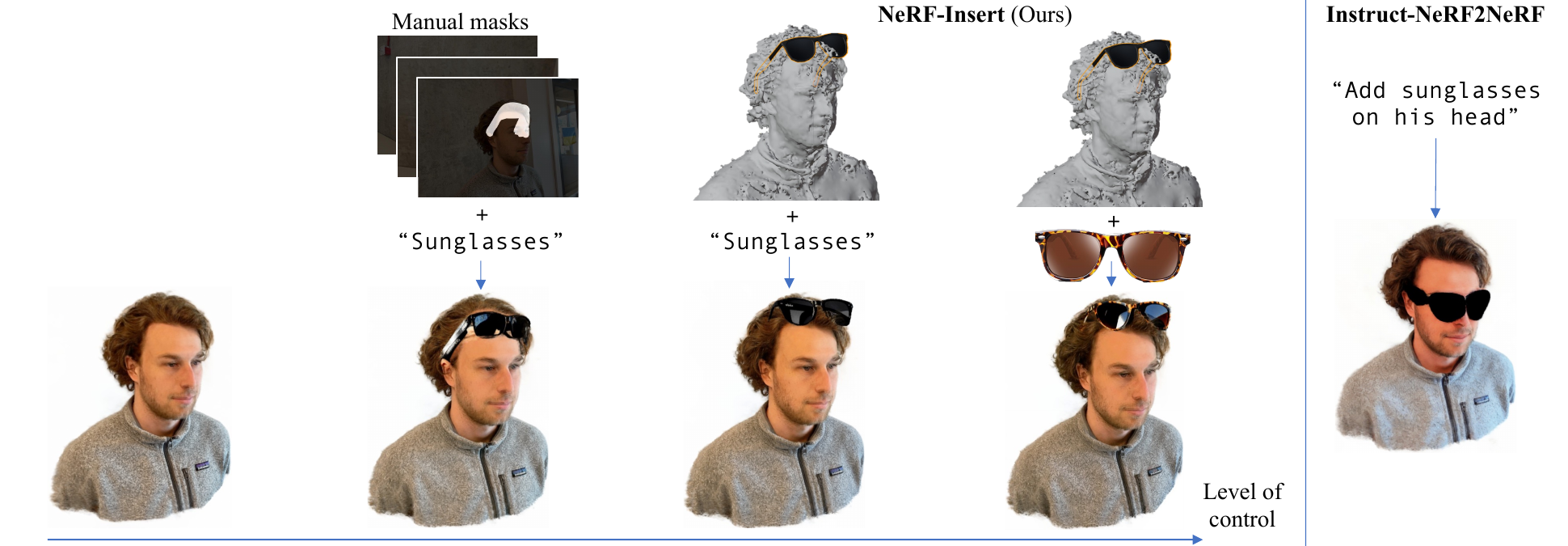} 
\caption{NeRF-Insert accepts a variety of conditioning inputs, which can be seen as an spectrum of levels of control. For example, a user can specify: 1) a textual description of an object and a rough 3D region where it should be inserted (via image masks); 2) a textual description of the object with its shape and pose determined by a CAD model; 3) additionally influence the appearance of the object via a reference image.
In contrast, text-based editing methods such as Instruct-NeRF2NeRF do not afford the same flexibility.
Our framework is generic enough to potentially incorporate other kinds of inpainting control modalities, for example masks from a segmentation model.}
\label{fig:control}
\end{figure}

While 2D diffusion models are capable of generating and editing high quality images, editing 3D scenes remains a challenge. Recently, image-to-image generative models have started being used to modify NeRFs with textual prompts. For example, Instruct-NeRF2NeRF \cite{haque2023instruct} uses image generation models \cite{brooks2023instructpix2pix} to edit individual views and lift the results to a 3D model. 
Methods such as Instruct-NeRF2NeRF, however, have difficulties applying \emph{local edits} to a scene: since image-to-image models can modify the entire input image, changes to the NeRF can impact the global structure of the scene in undesired and unpredictable ways (Figure~\ref{fig:overall}, right).
Moreover, the image generation methods used for existing NeRF editing methods typically consider only textual prompts as conditioning, thus providing very limited control over the editing process. In an ideal setting, a user should have the possibility to decide the extent to which the edits to the NeRF are controlled or constrained.


In this paper, we introduce NeRF-Insert, a framework for NeRF editing which inpaints only specific 3D regions of a scene and accepts different types of conditioning inputs with varying of levels of control. In particular, a user can provide a combination of textual prompts and reference images describing the edits, and specify a 3D region where the edits will be applied. The region can be described by either using a small number of manually drawn masks (typically 2-3 suffice) or by positioning a CAD model within the scene. In the latter case, the location and orientation of the inserted object is strongly constrained.
In addition to allowing different levels of control, we show empirically that our approach results in higher quality edits than previous methods and maintains stronger consistency with the original NeRF. 

To achieve this, NeRF-Insert uses a series of simple key ideas that could be applied more generally on top of any existing image editing model to lift 2D edits to a NeRF model. We use a visual hull to represent the region of space to be edited, which can be specified using a small number of masks in different views. In addition to using text-guided diffusion models, we also apply PaintByExample \cite{yang2023paintbyexample} to inpaint scenes with a reference image that conditions the appearance of the inserted object or edited region. Finally, we show how our framework can use a reference mesh/CAD model to constrain the geometry and pose of an inpainted object in a more fine-grained manner, as an alternative to manually drawing masks. Figure~\ref{fig:control} shows examples of how a scene can be edited with varying levels of control.

\section{Related Work}
\paragraph{Text-to-3D.} 
3D content generation is limited by the current limited scale of 3D datasets. Recent work tries to bypass this issue by leveraging visual priors learned by 2D diffusion models on large-scale image datasets. The main challenge with this approach is achieving 3D consistency of the images generated by stochastic 2D diffusion models. Dreamfusion \cite{poole2022dreamfusion} proposes Score Distillation Sampling (SDS) loss, which allows to achieve a 3D consistent asset represented by a NeRF. We discuss the SDS loss in more detail in section \ref{sec:sds}. Recent work on text-to-3D methods build upon the SDS loss \cite{Lin2023Magic3D, chen2023fantasia3d, zhu2023hifa}. 
Multy-view diffusion models \cite{shi2023mvdream, liu2023zero, shi2023zero123++} use a pose-conditioned diffusion model to synthesize novel views of an object, which are then used to generate high fidelity 3D assets. While this approaches have shown promising results in generating single objects, they still fail at generating complex 3D scenes, since these models are mainly trained on simple 3D assets and have not yet shown success in larger scenes as 2D diffusion models have.

\paragraph{Editing NeRFs.} Objects in a NeRF can be disentangled by segmenting the 3D space and modifying the field according to a rigid body transformation of that region \cite{yang2021learning, kundu2022panoptic, cen2023segment, Ost2021Neural}. However, this limits the edits to moving, scaling and rotating existing objects. \cite{liu2021editing} allows for editing from user scribbles, but is limited to a per-category conditional radiance field (i.e. chairs) and limited scribble modifications such as removing or changing the color of some parts of the NeRF. Clip-NERF \cite{wang2022clipnerf} edits conditional radiance fields by learning a CLIP-guided appearance and deformation network. However, it is limited to simple modifications and requires training a conditional model from scratch. \cite{yuan2022nerfediting} edits the scene by extracting a mesh, editing its geometry and transferring the mesh edits back into the implicit representation, but is also limited to geometric edits and requires the use of an external mesh editing tool. More similar to our method, Instruct-Nerf2Nerf \cite{haque2023instruct} leverages a pre-trained diffusion model and adopts an Iterative Dataset Update (IDU) method to distill the edits in a 3D consistent way into the pre-trained existing NeRF scene. As we elaborate in section \ref{sec:results}, this model fails at localized editing, sometimes modifying the global appearance and structure of the scene and inpainting, inserting blurry or noisy objects, or not inserting anything at all. These artifacts are due to limitations of the 2D editing model, or due to difficulties in achieving 3D consistency from non-consistent 3D edits. 
Recent work \cite{yu2023editdiffnerf, fang2023text, pang2023locally, park2023ed, wang2023proteusnerf} improves over Instruct-NeRF2NeRF but edits remain global, often low quality and suffering from similar kinds of artifacts, specially when trying to add non-existing objects. Focaldreamer \cite{li2023focaldreamer} allows to locally edit a 3D asset via text prompts and intuitive region query but is limited so single objects and has not shown good results in real-world complex scenes. 
Diffusion models have also been used to inpaint real world, complex 3D scenes \cite{mirzaei2023spinnerf, wang2023inpaintnerf360, weder2023removing} but previous work is limited to removing existing objects in the scene and have not shown success in text-guided inpainting. 
DreamEditor \cite{zhuang2023dreameditor} allows editing 3D scenes with prompts but does not allow for specifying the spatial 3D location with intuitive modalities such as hand-drawn coarse masks. This leads to similar problems to those encountered with Instruct-NeRF2NeRF, where the edited region needs to be clearly defined by a text prompt alone. NeRF-Insert, on the other hand, enables more precise control on the inpainting region and pose of the inserted object.
Vox-E \cite{sella2023vox} enables local editing of 3D scenes with text prompts but shares the same kind of limitations as Dreameditor and moreover does not perform well on large and complex scenes.
Finally, we mention TIP-Editor \cite{zhuang2024tip}, which is concurrent to our work and also considers the problem 3D scene editing using different control modalities. However, unlike our method, it restricts spatial location specification to 3D bounding boxes, limiting the ability to define editing regions of arbitrary shape.

\paragraph{Mask 2D-to-3D lifting.} Prior work lifts 2D masks or user clicks into a 3D mask or region that can be used for editing such as rigid body transformation \cite{cen2023segment} or inpainting \cite{wang2023inpaintnerf360, mirzaei2023spinnerf}. These methods project the 2D mask into a 3D region by leveraging known 3D scene depth and use a pre-trained image segmentation model \cite{kirillov2023SAM} to complete the mask from missing view points. The main limitation of this approach is that segmentation models can only segment already existing elements in a scene, and prevents the user from defining a region that does not correspond to an existing element, which is a paramount property when we want to edit arbitrary parts of the scene or insert non-existing objects.

\section{Preliminaries}
\paragraph{NeRFs.}
Neural Radiance Fields (NeRF) \cite{mildenhall2021nerf} have become a commonly used 3D representation for view interpolation\cite{mildenhall2021nerf, barron2021mip, barron2023zip}, given their ability to represent complex scenes in a very compact way. A scene can be parametrized by a positional hash encoding followed by small a multilayer perceptron \cite{muller2022instant}, which encode the scalar density $\sigma(\boldsymbol{x})$ and the three-dimensional (RGB) radiance $\boldsymbol{c}(\boldsymbol{x}, \theta, \tau)$ of a field, being $\boldsymbol{x}$ the 3D position of space and $\theta, \tau$ the view direction. Arbitrary viewpoints can be rendered via volume rendering and the model's parameters are optimized via RGB supervision of posed images. The volume rendering equation 
defines the color $C(\boldsymbol{r})$ of the pixel defined by ray $\boldsymbol{r}$. $N$ positions are sampled across the ray, and the radiance field is then queried in these positions, providing the density $\sigma_i$ and color $\boldsymbol{c}_i$ of each sample $i$ for $0<i\leq N$. \cite{barron2021mip, barron2023zip} extend this by volume rendering frustums as opposed to single rays, and is implemented by the used framework Nerfstudio\cite{tancik2023nerfstudio}. $C(\boldsymbol{r}) = \sum_{i=1}^N w_i \boldsymbol{c}_i$, being $w_i = \Psi(\sigma_i \ \delta_i) \prod_{j=1}^{i-1} (1 -\Psi(\sigma_j \delta_j))$, where $\Psi(x) = 1 - e^{-x}$
 and 
$\sigma_i$ is the density of $i$th sample,  $\delta_i$ the distance between the $i$th and $i-1$ sample.

\paragraph{Inpainting diffusion models.} Image inpainting consists in filling in missing or masked parts of an image. Diffusion-based models have enabled high quality image generation and have been adapted for high quality image inpainting. Diffusion models gradually denoise a random vector $z_T$ into a valid image \cite{ho2020DDPM} or a latent vector $z_0$ which is then decoded into a valid image \cite{song2020DDIM, rombach2022StableDiffusion, razzhigaev2023kandinsky, zhang2023controlnet}. This method simulates a Markov chain for multiple steps, each corresponding to a gradually less \textit{noisy} version of the \textit{denoised} image on the final step. 
During inference, efficient denoising algorithms such as \cite{song2020DDIM} approximate the full Markov chain with a reduced number of deonising steps, and can be adopted to balance computational costs and sample quality. Text-to-image diffusion models condition the gradual denoising steps to a CLIP embedding \cite{radford2021CLIPlearning} of a text prompt which allows to control the content of the generated images. Inpainting models adopt a similar strategy to image generation methods, but after each denoising step replace the predicted denoised image outside the mask by a noisy version of the original image, effectively keeping the region outside of the mask as the original image and generating new content for the region inside the mask.
Stable diffusion \cite{rombach2022StableDiffusion} is a widely used base model for text-to-image generation that has also been adapted for inpainting. 
Paint-By-Example \cite{yang2023paintbyexample} is a diffusion-based image inpainting model that allows to condition the inpainting to an additional reference image which replaces a textual prompt. 

\paragraph{Score Distillation Sampling.}
\label{sec:sds}
Score Distillation Sampling (SDS) is a loss term introduced by Dreamfusion \cite{poole2022dreamfusion} that allows to generate 3D assets with 2D diffusion models.
During the forward pass of a diffusion model, it is possible to generate a sample from a noisy version of a reference image, as opposed to from a purely noisy sample. Starting from a purely noisy sample allows to cover a wider, more general distribution and starting from a less noisy version of a reference image will generate a sample that resembles more the reference image. This is the main property that initially allowed diffusion models to be used to generate view-consistent 3D assets. Score Distillation Sampling is the loss term proposed by Dreamfusion \cite{poole2022dreamfusion} which consists in iteratively denoising the NeRF renders from different levels of noise and optimizing the NeRF by supervising it with the denoised images, making the radiance field converge into a 3D-consistent representation that follows the diffusion prior. Different variations and improvements have been proposed for this loss term, such as Iterative Dataset Update (IDU)\cite{haque2023instruct}, or annealed noise scheduling \cite{zhu2023hifa, shi2023mvdream}. To the extent of our knowledge, no prior work has shown effectiveness of the Score Distillation Sampling approach with text-conditioned inpainting models.

\section{NeRF-Insert}

\begin{figure}[h]
\centering
\includegraphics[width=0.99\linewidth]{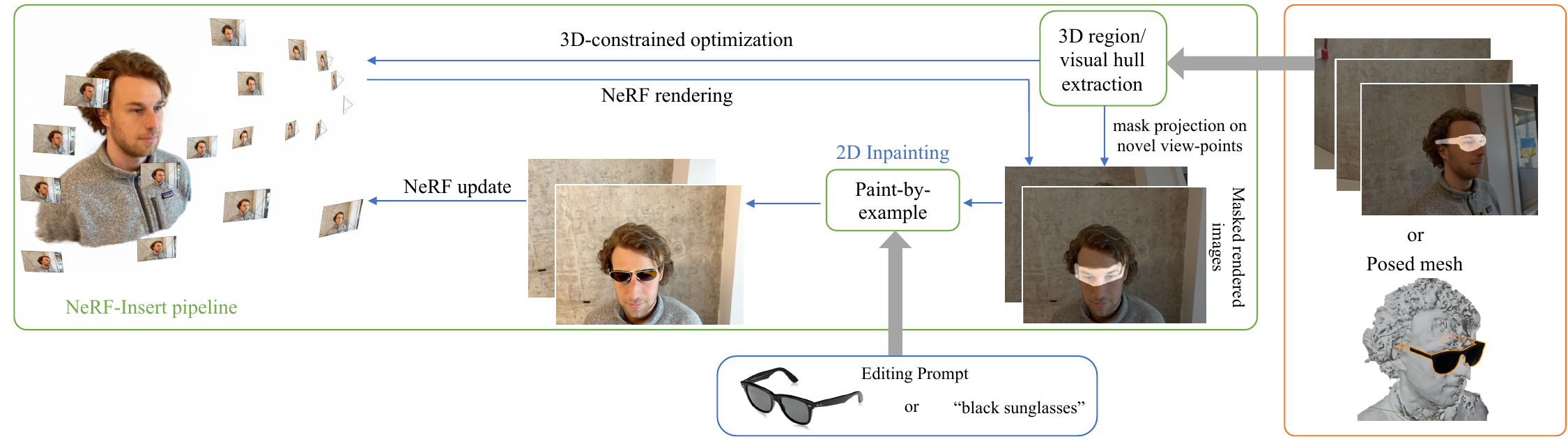} 
\caption{Overview of NeRF-Insert. We use a small number of manually-drawn masks or a posed mesh to define the 3D region of space to edit. This 3D region is projected onto the training views to obtain inpainting masks fo all of the training images. We render the NeRF from the training viewpoints and inpaint them using Stable Diffusion or Paint-by-Example. We then replace the previous images in the training pipeline of the NeRF with the inpainted images.}
\label{fig:model_diagram}
\end{figure}

Our editing method is based on an Iterative Dataset Update, which iteratively replaces the training images for edited (inpainted) images. Our method can incorporate different inpainting models, namely text-guided and image-guided models. In order to have a well-defined inpainting mask from all training views, we propose different ways for the user to intuitively define an inpainting 3D region which can then be rendered from the training views to generate the masks needed for inpainting. This 3D region can be defined by as few as 2 manually drawn masks or by placing a mesh in the scene. Finally, we propose an additional loss term that enforces the edits to be made within a specific region of the 3D space, which we show significantly improves on the quality of the edited 3D scene. Figure \ref{fig:model_diagram} presents a high-level overview of NeRF-Insert.

\subsection{Distilling 2D inpainting into 3D}
Following \cite{haque2023instruct} we adopt an Iterative Dataset Update algorithm to distill the 2D edits made by the inpainting model into the NeRF 3D representation. This algorithm consists of four steps. First, the NeRF is rendered from the training views. Second, we use the inpainting diffusion models to inpaint a noisy version of the rendered training images.
As described in Section \ref{sec:sds}, adding more initial noise to the rendered training images will give the inpainting model more freedom to inpaint the masked region while adding less noise will preserve more similarity with the rendered images, posing a trade-off between editing capabilities and consistency with the current state of the NeRF. 
Third, the training images are replaced by the corresponding inpainted images. Finally, the NeRF continues to be trained for a certain number of iterations before the process is repeated.
This iterative procedure eventually converges into a 3D consistent representation that is guided by the diffusion model.

\paragraph{Noise annealing.} Following \cite{zhu2023hifa, shi2023mvdream} we adopt an annealed noise scheduling over the course of the optimization procedure. In other words, at the beginning of the training procedure we denoise a purely noisy latent, which makes the output of the diffusion model very inconsistent across views and makes the NeRF converge into a blurry scene. As the training proceeds, the next rounds of inpainting will be generated from a progressively less and less noisy version of the render of the current state of the NeRF. This decreasing noise schedule has shown to help convergence and greatly improve sample quality in text-to-3D methods with respect to using a random noise level at each iteration  \cite{zhu2023hifa}. In order to modulate the amount of initial noise added to the sample before the diffusion denoising, we vary the \textit{strength} $s$ of the diffusion process. The \textit{strength} refers to the amount of initial noise added and the corresponding number of denoising diffusion steps, as defined by the inference denoising scheduler. $s=1$ denoises a purely noisy vector \textit{from scratch} and $s=0$ adds no initial noise at all. We schedule $s$ at training step $n$ as: $s_n = 1.0-0.8\sqrt{\frac{n}{N_{steps}}}$, gradually decreasing the strength from $1.0$ to $0.2$ throughout a total of $N_{steps}$ NeRF training steps.

We update the \textit{whole} dataset every $N_{update}$ NeRF training iterations, unlike \cite{haque2023instruct} which edits a single training view every 10 iterations. This allows to batch diffusion forward passes, making it more efficient, and we observed it allows for a better convergence into a 3D consistent representation.

\subsection{Visual hull field}
\label{sec:visualhull}
We propose to use a binary field represented by a visual hull \cite{laurentini1994visualhull} that allows to define a 3D region from few (typically 2,3) user-generated masks or silhouettes. A visual hull is a technique from classical 3D computer vision that is used to perform shape-from-silhouette. Given the silhouette of an object seen from multiple cameras with known pose, the visual hull is the region of the 3D space where the frustums generated by each of the silhouettes intersect. Similarly, we use a small set of user-generated masks to define a 3D region consisting of the intersection of volumes generated by each of the masks. 

We implement the visual hull by projecting a 3D point onto the views of the manually drawn masks and checking if the projected points fall within all the respective masks. An additional property of this implementation is that it is very efficient to test if an arbitrary point in space is inside or outside the visual hull, which makes it suitable to query millions of time per second.

Alternatively, the manual masks can be replaced by using a reference mesh of the object that we want to insert. We first place the object in the location where we want the edit to happen. We then render that object from multiple views using a solid white texture and black background, which effectively serves the purpose of manually drawing the masks from different view-points, but with a higher level of precision in the region being defined.

\subsection{Mask reprojection}

\begin{figure}[h]
     \centering
     \begin{subfigure}[b]{0.49\textwidth}
         \includegraphics[width=\textwidth]{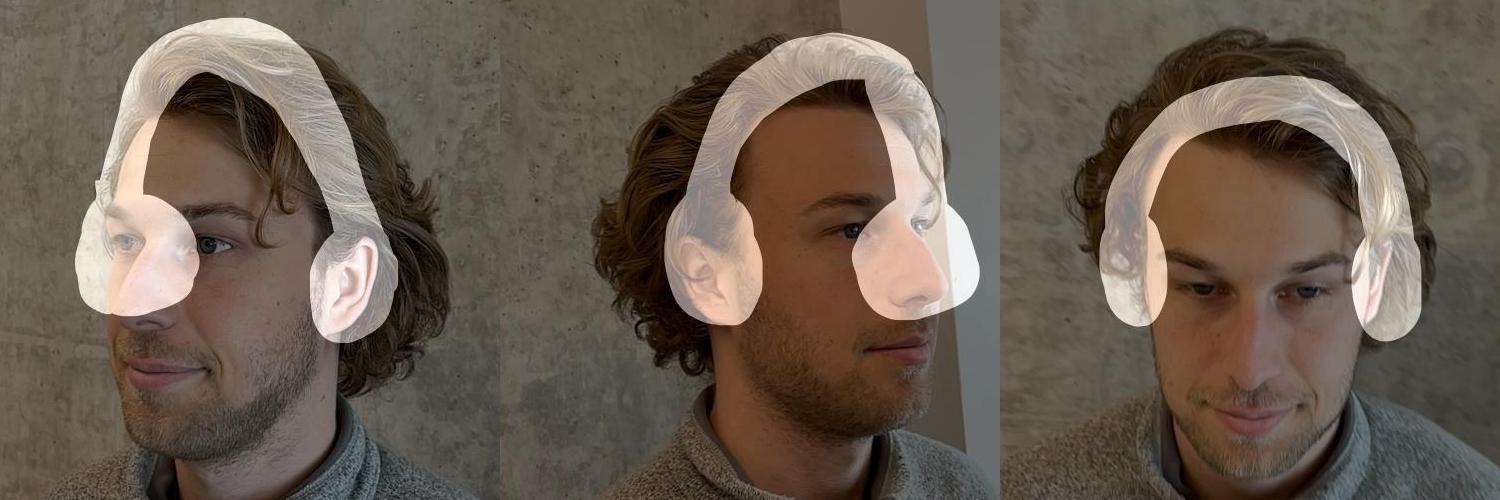}
         \caption{Three manually annotated masks.}
         \label{fig:hp_manualmasks}
     \end{subfigure}
     \begin{subfigure}[b]{0.49\textwidth}
         \includegraphics[width=\textwidth]{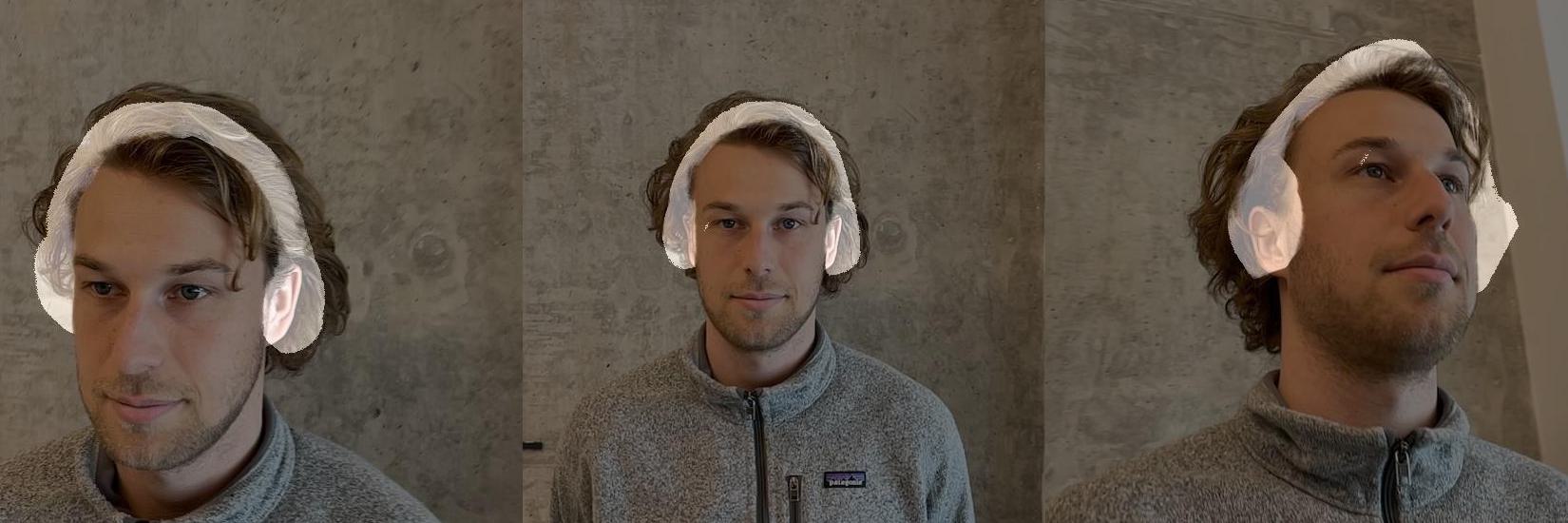}
        \caption{Mask reprojection onto other views.}
        \label{fig:lifted}
     \end{subfigure}
     \caption{We lift three manually-drawn masks to a 3D representation that can be rendered from an arbitrary viewpoint. As we see in the leftmost and rightmost masks in the bottom row, our projection accounts for occlusions in the existing scene.}
     \label{fig:hp_masklift}
\end{figure}

In order to inpaint the scene following the iterative dataset update, we need to inpaint every training image, and to do so we need an inpainting mask for each corresponding viewpoint. Manually drawing the inpainting region for every training view would be time-consuming, so we use the mask field defined by a very small amount of manually-drawn masks to render inpainting masks from all view-points. More specifically, we render the visual hull from any viewpoint using volume rendering, which can be naturally incorporated into a NeRF pipeline. In order to do so, we first define a binary radiance field which has high density value and white radiance \textit{inside} the visual hull and has zero density and black radiance \textit{outside} the visual hull. Volume rendering this field with would generate the silhouette of the visual hull from an arbitrary view point. Simply rendering the silhouette would not take into account occlusions with the unedited scene. In order to do so, instead of setting the density outside of the visual hull to zero, we keep the same density in original NeRF field. Figure \ref{fig:hp_masklift} shows an example of 3 manually drawn masks, together with the rendering of the visual hull from 3 different views. As we can observe, the visual hull can be rendered from different viewpoints while properly accounting for occlusions with the existing scene.

\subsection{Spatially constrained 3D edit}
NeRFs are optimized via 2D supervision. Our method is optimized using the loss in Nerfacto and Instruct-NeRF2NeRF, which is a weighted sum of the RGB loss, LPIPS loss \cite{zhang2018perceptual}, and distortion loss \cite{barron2022mip}. 
In NeRF-Insert, we want to edit exclusively inside a 3D region and ensure the outside of that region remains the same as the original NeRF. Because of the implicit nature of the NeRF representation, it is not possible to \textit{freeze} certain regions of space, since all spatial locations are entangled in the model's parameters.
We propose to spatially constrain the modifications on the scene by adding an additional loss term. In order to do this, we keep a copy of the unedited NeRF, and add a mean square error loss term to enforce that the 3D points sampled across each ray that fall outside the 3D mask have the same density and color than the original unedited NeRF. As shown in Equation \ref{eq:lossterm}, we scale this loss term at each point by its contribution to the color of the pixel (often referred to as the \textit{weight} in the volume rendering equation). Scaling the loss term with this weight allows to \textit{not} enforce any field values in the regions that are transparent or occluded by other parts of the scene.

\begin{equation}
\begin{split}
        \mathcal{L}_{out} =
    \frac{1}{M}
    \sum_{i \in ray}
    (
    &(\lambda_c||\textbf{c}_i - \hat{\textbf{c}}||^2 + \lambda_{\sigma}|\Psi(\sigma_i) - \Psi(\hat{\sigma_i})|^2) \\ & * (\hat{w}_i + w_i) * (1 - m_i)
    )
\end{split}
\label{eq:lossterm}
\end{equation}

Here $\hat{w}_i$ 
 is the weight attributed to the color in the volume rendering equation (visibility factor), $m_i \in \{0,1\}$ is the binary mask correspondence of the 3D point, 
 and $M = \sum_{i \in ray}(1-m_i)$. The reason why we add the visibility factor from the unedited NeRF the edited one is to ensure that we apply this loss term to any 3D point that is visible from the edited \textit{or} the unedited NeRF.

 \section{Results}

\begin{figure*}[h]
\includegraphics[width=0.99\linewidth]{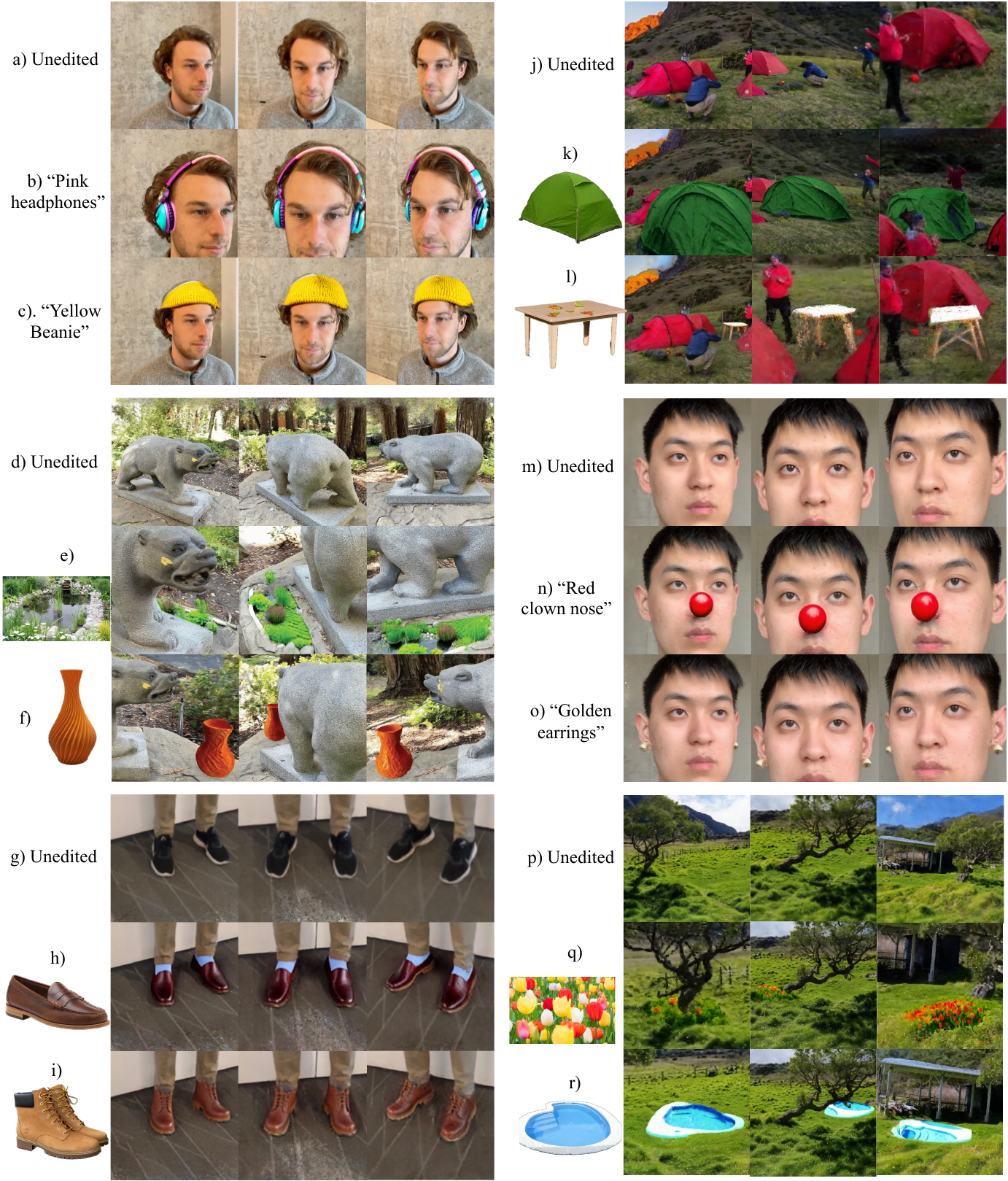} 
\caption{Inpainting results. On the left side there is the textual or visual prompts used for Stable Diffusion inpainting or Paint-by-Example, respectively. Additionally, to specify the inpainting 3D region, b), c), e), h), i), n) rely on 3 manually drawn masks, f) relies on a geometrically accurate mesh of a vase while k), l) o), q), r) rely on geometrically coarse meshes such as a cube, sphere or cylinder. Refer to the supplementary material for more details about masks and meshes used for each example.}
\label{fig:examples}
\end{figure*}

\begin{figure}[h]
\centering
\includegraphics[width=0.99\linewidth]{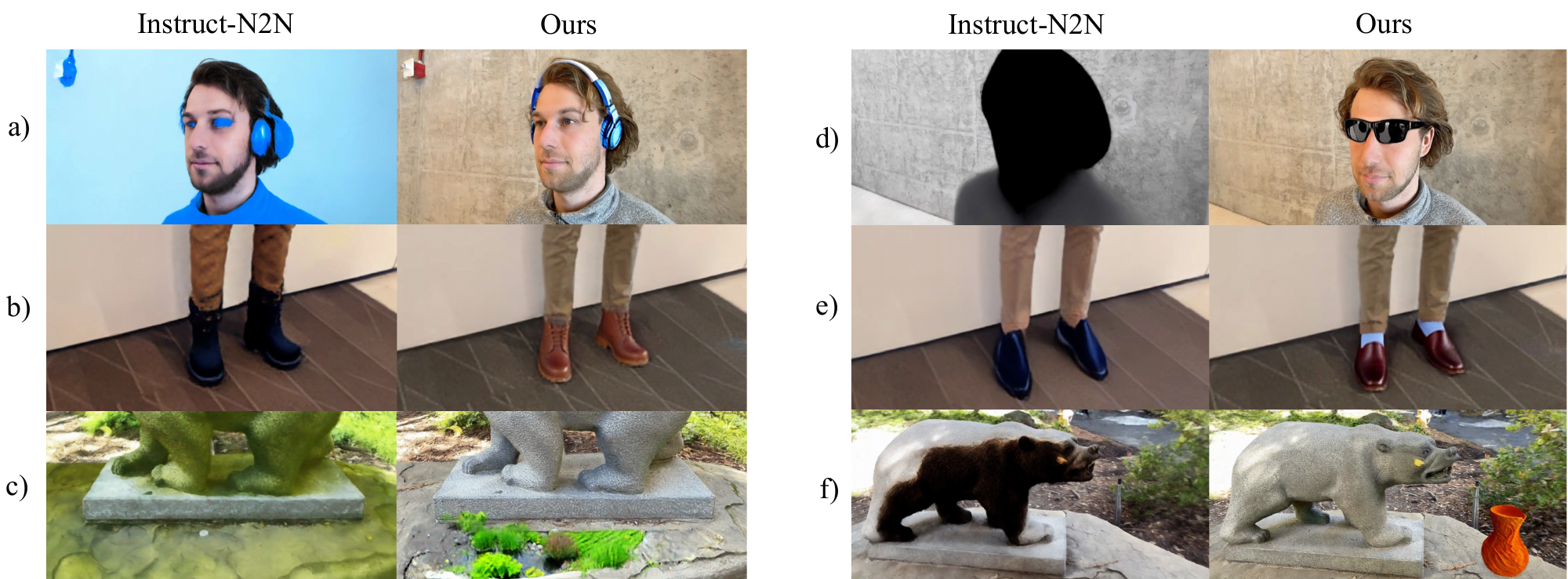} 
\caption{Comparison of Instruct-NeRF2NeRF (\emph{left}) and NeRF-Insert (\emph{right}). I-N2N is not able to constrain the edit according to the given prompt and the whole structure of the scene is modified. Prompts for Instruct-NeRF2Nerf are [\textit{Give him}] a) \textit{blue headphones}; d) \textit{black sunglasses}; b) \textit{boots}; e) \textit{loafers}; and [\textit{Add}] c) \textit{a garden pond next to the bear}; f) \textit{a vase in front of the bear}. Inpainting prompts and inpainting regions are defined in  Figure~\ref{fig:examples} and supplementary material..}
\label{fig:in2n}
\end{figure}

\begin{figure}[h]
\centering
\includegraphics[width=0.99\linewidth]{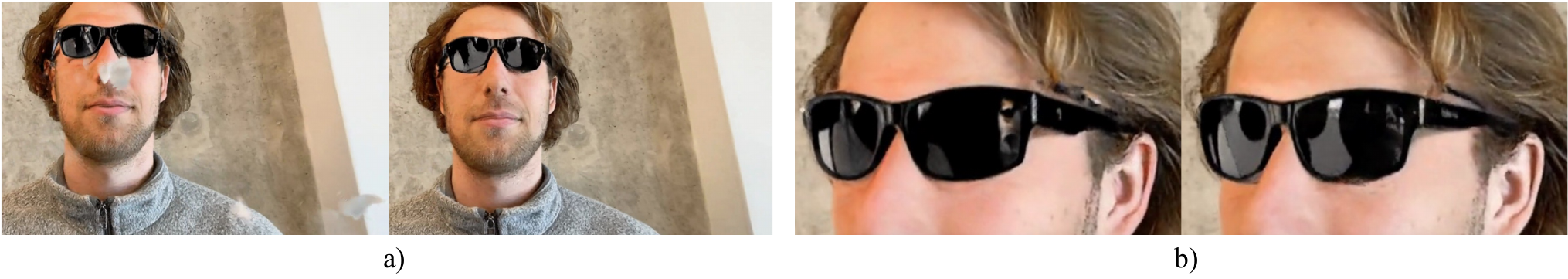} 
\caption{Results without and with (\emph{right}) the additional loss term~\eqref{eq:lossterm}. Using the in-mask edit loss term eliminates floaters outside the mask artifacts (a) and increases quality of the inpainting in the mask (b). The artifacts in (b) are hard to appreciate in still images but become much more apparent in the videos provided in the supplementary material.}
\label{fig:outside_mask}
\end{figure}

\label{sec:results}
We show qualitative and quantitative results for multiple real scenes, which were openly released by Instruct-NeRF2NeRF \cite{haque2023instruct}. We show how previous work fails at local editing while NeRF-Insert is capable of high-quality inpainting in a variety of real scenes, with different prompts and in different regions of the scene. We also show the impact of the constrained edit loss term, which removes floating artifacts and significantly improves the quality of the edited region. Our results are best viewed in the videos included in the supplementary material.

\paragraph{Implementation details.}
We use the Nerfacto model from Nerfstudio \cite{tancik2023nerfstudio} as our backbone representation, and we remove the view direction dependency on the color field, which improved view-consistency while not impacting much the quality of the NeRF for our scenes. For text-prompted and image-prompted inpainting, we use Stable Diffusion \footnote{\url{https://huggingface.co/stabilityai/stable-diffusion-2-inpainting}} and Paint-by-Example\footnote{\url{https://huggingface.co/docs/diffusers/api/pipelines/paint_by_example}} models respectively, publicly available in the Huggingface Diffusers library. For the text-based diffusion, we append "Man wearing" to prompts like "blue headphones" or "sunglasses". 

The Iterative Dataset Update algorithm updates the whole set of training images every 6000 NeRF training steps and will do it over the course of $90,000$ steps. The procedure takes around 2 hours, depending on the scene. Before inpainting, we crop the images around the mask region in order to ensure that the inpainting mask is large enough for the inpainting model to function properly. Randomly sampling the crop size shows quality improvement in some scenes such as the shoes. Crop size intervals depend on the scene and inpaint mask and are reported in the supplementary material. For inpainting, we dilate the mask with a round kernel of $11$ pixels of diameter and after inpainting we use the non-dilated mask to crop the inpainting back into the unedited image. In the constrained loss term we use $\lambda_c=100$ and $\lambda_\sigma=1000$ and add it to the default Nerfacto loss.


\paragraph{Inpainting NeRF Scenes.}
Nerf-Insert is able to inpaint a variety of scenes and prompts. Our method is able to add objects or modify the scene at specific 3D region and ensures the scene is not modified outside of it. In Figure~\ref{fig:examples} we show a variety of examples from three different real scenes which we inpaint in different regions with different guidance prompts.
In all these scenes we demonstrate how NeRF-Insert is not only able to add an object in a specific region but also to edit the scene in order to accommodate the object being inserted. This is analogous to how 2D inpainting models differ from a \textit{crop and paste} approach. We see how the different objects wrap the head properly and integrate well with the hair. Similarly, the garden pond is nicely integrated into the rock and the shoes fit the legs properly. Apart from experiments that use a mesh to define the inpainting region, the edits in the face scene rely on three manually drawn masks and the edits in the bear and shoes scenes use two manually drawn masks. We provide visualization of the manual masks and reprojected masks in the supplementary material. In Figure \ref{fig:in2n} we compare our results with Instruct-NeRF2NeRF. Our method is able to constrain the edits in a single region while Instruct-NeRF2NeRF often modifies the whole scene or generates poor quality edits. 

We observed how image-guided inapainting with Paint-by-Example often outperforms Stable Diffusion in complex prompts. We hypothesize that this is partially due to the fact the textual descriptions are more visually ambiguous and hence it is harder to converge to a specific 3D-consistent representation. We also hypothesize that using other state-of-the-art inpainting models such as \cite{razzhigaev2023kandinsky, tang2023realfill} could increase the performance of the overall method.

Using a reference mesh instead of using manual masks, as in Figure~\ref{fig:examples}(i) (sunglasses) or (o) (vase), allows the user to specify the inpaint region with higher level of precision. As seen in Figure \ref{fig:control}, using a mesh only constrains the orientation, position and geometry of the object, but the appearance can still be controlled independently as we see in the example of the leopard sunglasses.

\paragraph{Quantitative results}. Editing is a subjective task, but following previous work \cite{haque2023instruct}, we report two quantitative CLIP-based metrics. The \textit{text edit direction} is defined as the difference or direction between the CLIP embedding of a generic text string describing the unedited scene such as "a man" and the CLIP embedding of a text string describing the edited scene such as "a man wearing blue headphones". Similarly, the \textit{image edit direction} is defined as the difference between the CLIP embedding of a frame of the original scene and the CLIP embedding of its corresponding frame from the edited scene. The CLIP Text-Image Direction Similarity measures the alignment between editing prompt and edited images by computing the average cosine similarity between text edit direction and image edit direction across frames. The CLIP Direction Consistency measures the editing consistency by computing the average cosine similarity between image edit directions of contiguous frames in a given rendered video. Following the same practice than Instruct-NeRF2NeRF, we compute these two metrics in a small set of 10 scene-prompt combinations (Figure \ref{tab:quant}). We see that our method produces scenes that are 110\% better aligned with the editing prompt than IN2N, at the cost of a modest decrease (4\%) in inter-frame consistency. Note that these metrics have some limitations: firstly, they do not measure the fact that IN2N edits are not localized, one of our key improvements. Moreover, invalid global edits (e.g., turning the whole scene blue as IN2N in Fig.~6a) can in fact increase consistency across frames, which might explain the small decrease in CLIP Direction Consistency.
\begin{table}
  \centering
  \resizebox{.5\columnwidth}{!}{
  \begin{tabular}{lc@{}cc@{}cc@{}}
    \toprule
    Method & CLIP Text-Image & CLIP Direction \\
    &Direction Similarity & Consistency  \\
    \midrule
    IN2N & 6.47 & \textbf{91.82} \\
    Ours & \textbf{13.62} & 88.05 \\
    \bottomrule
\end{tabular}}
    \caption{Quantitative metrics on 10 scene-prompt combinations.}
  \label{tab:quant}
  \vspace{-.5cm}
\end{table}

\paragraph{Analysis.} Our mask reprojection method determines an inpainting mask in any arbitrary viewpoint, given a visual hull generated by as few as 2 manually drawn masks, while accounting for occlusions in the existing scene.
Not accounting for occlusions with the scene would generate an inpainting mask which would include part of the scene that should be preserved, for example the nose and eyes from the left and middle view of Figure \ref{fig:hp_manualmasks}. We show how the visual hull is well suited not only for mask reprojection but also to define an arbitrarily-shaped 3D region of space in which to implicitly constrain the modifications on the NeRF.

Our proposed loss term if effective at constraining the modifications to a region of 3D space and keeping the rest of the scene untouched.
If the editing is not constrained in space, we observe floaters appearing between the camera and the region we want to edit. Moreover, these floaters impact the quality of the region we want to inpaint, creating transparent and noisy areas in the inserted object. In figure \ref{fig:outside_mask} we show how using our additional loss term removes floaters and increases the quality of the generated sunglasses. These results become much more apparent in the videos included in the supplementary material.

\clearpage
\section{Conclusions}

In this paper, we presented NeRF-Insert, a method to locally edit a NeRF using a combination of textual and visual conditioning inputs. Our method allows the user to decide how strongly to control the editing process. For example, fine-grained control can be obtained by using a mesh to determine the inserted object's shape and pose, or using a reference image to influence its appearance. We note that NeRF-Insert suffers from artifacts similar to early SDS-based text-to-3D models. These limitations include converging to a noisy or inconsistent edit or potentially having the multi-face Janus problem. These problems might be diminished by using multi-view diffusion models trained on complex scenes. Another limitation of our method is that manually drawing a masks can be difficult without a proper interface and using a mesh or CAD to define the region is not always possible. 

Overall, NeRF-Insert is a flexible framework that could accommodate different kinds of control signals and inpainting models. 
For example, models such as ControlNet \cite{zhang2023controlnet} could be used to control the edits in different ways such as manual sketches or segmentation masks. Similarly, other methods such as SegmentAnything \cite{kirillov2023SAM, cen2023segment} could be applied to generate inpainting regions. We leave these directions for future work.

\clearpage
\bibliographystyle{splncs04}
\bibliography{main}
\end{document}